\pdfoutput=1

\documentclass[11pt]{article}

\usepackage{ACL2023}

\usepackage{times}
\usepackage{latexsym}

\usepackage[T1]{fontenc}

\usepackage[utf8]{inputenc}

\usepackage{microtype}

\usepackage{inconsolata}

\usepackage{multirow}
\usepackage{makecell}

\usepackage{times}
\usepackage{latexsym}
\usepackage{graphicx}
\usepackage{subfigure}
\usepackage{float}
\graphicspath{{images}}
\usepackage{stfloats}
\usepackage{booktabs}
\usepackage{indentfirst}
\usepackage{amsmath}
\usepackage{amssymb}

\usepackage{enumitem}

\usepackage{colortbl}
\usepackage{arydshln}

\usepackage[linesnumbered,ruled,vlined]{algorithm2e}

\SetCommentSty{mycommfont}

\SetKwInput{KwInput}{Input}                
\SetKwInput{KwOutput}{Output}              

\usepackage{CJKutf8}

\usepackage{listings}
\usepackage{xcolor}

\definecolor{codegreen}{rgb}{0,0.6,0}
\definecolor{codegray}{rgb}{0.5,0.5,0.5}
\definecolor{codepurple}{rgb}{0.58,0,0.82}
\definecolor{backcolour}{rgb}{0.95,0.95,0.92}

\lstdefinestyle{mystyle}{
    backgroundcolor=\color{backcolour},   
    commentstyle=\color{codegreen},
    keywordstyle=\color{magenta},
    numberstyle=\tiny\color{codegray},
    stringstyle=\color{codepurple},
    basicstyle=\ttfamily\small,
    breakatwhitespace=false,         
    breaklines=true,                 
    captionpos=b,                    
    keepspaces=true,                 
    numbers=left,                    
    numbersep=5pt,                  
    showspaces=false,                
    showstringspaces=false,
    showtabs=false,                  
    tabsize=2
}

\title{MiLoRA: Efficient \underline{Mi}xture of \underline{Lo}w-\underline{R}ank \underline{A}daptation for Large Language Models Fine-tuning}

\author{
Jingfan Zhang$^1$$^{*}$ \ \ 
Yi Zhao$^2$\thanks{\ \ Equal contributions. } \ \
Dan Chen$^3$$^{\dagger}$ \ \ 
Xing Tian$^4$  \ \ 
\textbf{Huanran Zheng}$^5$ \ \
\textbf{Wei Zhu}$^5$\thanks{\ \ Corresponding author. For any inquiries, please contact: michaelwzhu91@gmail.com; } \\ 
\textsuperscript{\rm 1} iFLYTEK Co., Ltd, China  \\
\textsuperscript{\rm 2} University of Pennsylvania, USA, \href{mailto:zhaoyi3@seas.upenn.edu}{zhaoyi3@seas.upenn.edu} \\
\textsuperscript{\rm 3} Lenovo Connect Co., Ltd, China\\
\textsuperscript{\rm 4} Niuxin Network Technology Co., Ltd, China  \\
\textsuperscript{\rm 5} East China Normal University, China \\
}

\begin{document}
\maketitle
\begin{abstract}

Low-rank adaptation (LoRA) and its mixture-of-experts (MOE) variants are highly effective parameter-efficient fine-tuning (PEFT) methods. However, they introduce significant latency in multi-tenant settings due to the LoRA modules and MOE routers added to multiple linear modules in the Transformer layer. To address this issue, we propose Mixture of Low-Rank Adaptation (MiLoRA), a novel and efficient LoRA variant. MiLoRA differs from previous MOE-style LoRA methods by considering each LoRA module as an expert and employing a prompt-aware routing mechanism. This mechanism calculates expert routing results once before generating the first new token and reuses these results for subsequent tokens, reducing latency. Extensive experiments and analysis on commonsense reasoning tasks, math reasoning tasks, and widely used LLM evaluation benchmarks demonstrate that MiLoRA consistently outperforms strong PEFT baselines with comparable tunable parameter budgets. Additionally, MiLoRA significantly reduces latency in multi-tenant settings compared to previous LoRA-based methods.

\end{abstract}

\begin{CJK*}{UTF8}{gbsn}

\section{Introduction}

Large language models (LLMs) have been achieving state-of-the-art (SOTA) results not only in various natural language processing tasks \cite{qin2023chatgpt,PromptCBLUE,text2dt_shared_task,Text2dt,zhu_etal_2021_paht,Li2023UnifiedDR,Zhu2023BADGESU,Zhang2023LECOIE,Zhu2023OverviewOT,guo-etal-2021-global,zhu-etal-2021-discovering,Zheng2023CandidateSF,info:doi/10.2196/17653,Zhang2023NAGNERAU,Zhang2023FastNERSU,Wang2023MultitaskEL,Zhu2019TheDS,Zhu2021LeeBERTLE,Zhang2021AutomaticSN,Wang2020MiningIH}, but also many challenging evaluation tasks \cite{huang2023c,li2023cmmlu,Cui2023UltraFeedbackBL,wang2024ts,yue2023-TCMEB} but also in numerous challenging evaluation tasks \cite{huang2023c,li2023cmmlu}, such as question answering, reasoning, math, safety, and instruction following. Although LLMs are evolving into general task solvers, fine-tuning remains essential for efficient LLM inference and for controlling the style of the generated content \cite{Xin2024ParameterEfficientFF,Ding2022DeltaTA}. Full-parameter fine-tuning of such large models is impractical due to the significant GPU memory and computational resources required. Consequently, parameter-efficient fine-tuning (PEFT) \cite{Zhang2023LearnedAA,2023arXiv230318223Z} has garnered considerable attention in the research community, as it typically involves tuning less than 1\% of the LLMs' parameters, thereby substantially reducing computational costs.

\begin{figure*}
\centering
\includegraphics[width=0.7\textwidth]{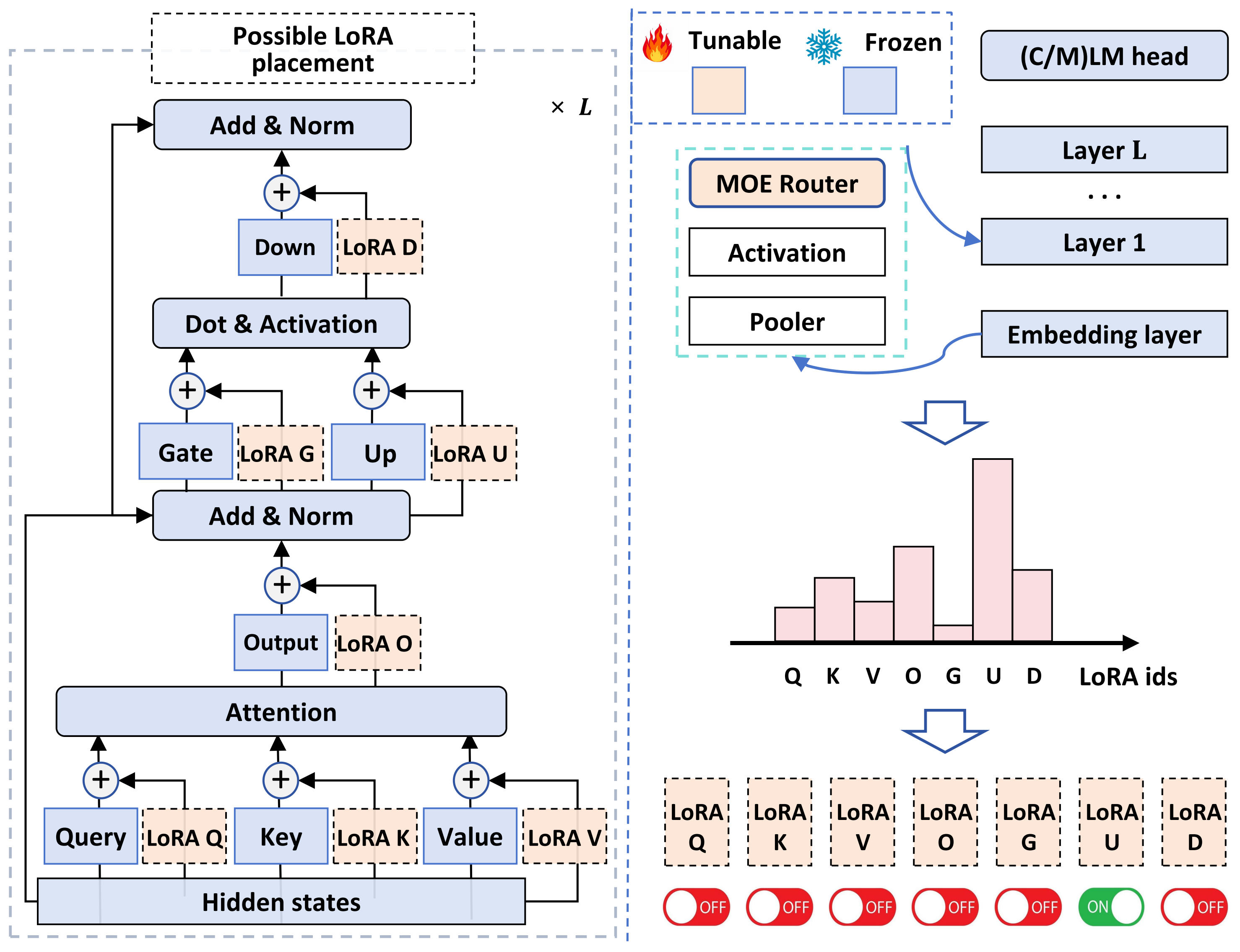}
\caption{Schematic illustration of our MiLoRA method. \textbf{Left}: The architecture of a Transformer layer as in LlaMA-2 \cite{Touvron2023Llama2O}. There are seven linear modules and seven positions to add LoRA modules. \textbf{Right}: Upon receiving an input prompt, the LoRA router before each Transformer layer will take the input prompt's hidden states as input features and go through a pooler, an activation function, and the MOE router network to determine which LoRA module is activated (or used) (e.g., LoRA U in the figure). This routing decision is repeatedly used when generating subsequent tokens. }
\label{fig:architecture}
\end{figure*}

Among many PEFT methods, the reparameterization-based method low-rank adaptation (LoRA) \cite{hu2021lora} is considered one of the most effective methods for LLMs \cite{Xu2023ParameterEfficientFM,Ding2022DeltaTA,Xin2024ParameterEfficientFF}. Although LoRA is effective and can bring stable performance with the original setting in \citet{hu2021lora}, it still brings inconvenience under the multi-tenant setting \cite{Chen2023PunicaML}: it has to add LoRA modules to multiple weights of the Transformer layer and introducing significant additional latency in every generation steps under the multi-tenant setting. Recently, the Mixture-of-Experts (MOE) style LoRA methods \cite{chen2024llava,yang2024moral,liu2023moelora,dou2023loramoe,gou2023mixture} have surged, further pushing the performance ceilings of LoRA fine-tuning. However, they introduce the calculation of MOE routers, further increasing inference latency. Thus, it is essential to develop a novel variant of the LoRA method that introduces minimum latency during generation and still can perform competitively in downstream tasks.

In this work, we propose a novel PEFT method called \underline{Mi}xture of \underline{Lo}w-\underline{R}ank \underline{A}daptation (MiLoRA). Our MiLoRA method differs from the previous literature on MOE-style LoRA methods in the following two aspects. First, in MiLoRA, an entire LoRA module is considered a LoRA expert, and the LoRA router is responsible for determining which LoRA expert to activate. Second, we propose the prompt-aware routing mechanism instead of calculating the expert routing results for every new token. Given an input prompt, the expert routing results are calculated once, right before the generation of the first new token. The subsequent generation steps will reuse the expert routing results. Under the prompt-aware routing mechanism, our LoRA router consists of a pooler operation, a learnable activation function \cite{Molina2019PadAU}, and a sparse MOE router.

We conduct extensive experiments and analysis on various challenging tasks, including five commonsense reasoning tasks, two math reasoning tasks, and three widely used LLM evaluation benchmarks. Our method can consistently outperform strong PEFT baselines with comparable tunable parameter budgets, especially the recent LoRA variants. In addition, our MiLoRA method has significantly lower latency under the multi-tenant setting \cite{Chen2023PunicaML} than the previous LoRA-based methods with comparable tunable parameters.

Our contributions are summarized as follows: 
\begin{itemize}
\item we propose a novel LoRA variant, MiLoRA, which combines the MOE mechanism with LoRA in an efficient way. 
\item In MiLoRA, we treat each LoRA module as an expert. 
\item We propose a prompt-aware routing mechanism to avoid token-wise router calculations. 
\item We have conducted extensive experiments and analysis showing that our MiLoRA framework is (a) practical and outperforms the baselines under comparable parameter budgets. (b) efficient during inference for LLMs. 
\end{itemize}

\section{Related works}

In the era of large language models, among the existing PEFT methods like Adapter \cite{Zhang2023LearnedAA,houlsby2019parameter}, Prompt tuning \cite{zhu2024iapt,zhu-tan-2023-spt} or (IA)$^3$ \cite{Liu2022FewShotPF,xie2024pedro} have been outperformed by LoRA \cite{liu2024alora,hu2021lora}. Since LoRA is the most popular PEFT method in the era of large language models~\cite{Cui2023UltraFeedbackBL,zheng2024nat4at,zhu2023acf,gao2023f,zuo-etal-2022-continually,zhang-etal-2022-pcee,sun-etal-2022-simple,zhu-etal-2021-gaml,Zhu2021MVPBERTMP,li-etal-2019-pingan,zhu2019panlp,zhu2019dr,zhou2019analysis}, many works are devoted to improving upon LoRA. AdaLoRA \cite{Zhang2023AdaptiveBA} looks into the parameter allocation of LoRA modules. VERA \cite{Kopiczko2023VeRAVR} investigate whether one could freeze the randomly initialized LoRA matrices and only learn a set of scaling vectors. Recently, a series of works has been looking into combining Mixture-of-Experts (MoE) \cite{shazeer2017outrageously,jacobs1991adaptive} and LoRA. LLaVA-MoLE \cite{chen2024llava} effectively routes tokens to domain-specific LoRA experts, mitigating data conflicts and achieving consistent performance gains over the original LoRA method. MOELoRA \cite{liu2023moelora} proves that fine-tuning LoRA modules with a MOE router enables the LLMs to perform well in a multi-task learning setting. MoRAL \cite{yang2024moral} addresses the challenge of adapting LLMs to new domains/tasks and enabling them to be efficient lifelong learners using the MOE techniques. LoRAMoE \cite{dou2023loramoe} integrates LoRAs using a router network to alleviate world
knowledge forgetting after instruction tuning. MoCLE \cite{gou2023mixture} proposes a MoE architecture to activate task-customized model parameters
based on instruction clusters. 

Although performing well in fine-tuning, these methods introduce high additional latency since (a) these methods do not reduce the number of LoRA modules in the Transformer backbone. (b) the routers and LoRA modules must be called when generating each new token. Our MiLoRA method addresses this efficiency issue by (a) only calling the LoRA routers when encoding the input prompt and before generating the first new token. (b) only activate one LoRA module per Transformer layer.

\section{Methods}

In this section, we first introduce the foundational concepts of LoRA and MoEs and then elaborate on the architectural design of MiLoRA.  

\subsection{Preliminaries}

\noindent \textbf{Transformer model} \quad As depicted in Figure \ref{fig:architecture}, each Transformer layer of a LLM such as LlaMA-2 \cite{Touvron2023Llama2O} consists of a multi-head self-attention (MHA) sub-layer and a fully connected feed-forward (FFN) sub-layer. MHA contains four linear modules, which are the Query (Q), Key (K), Value (V), and Output (O) modules. FFN contains three linear modules: Gate (G), Up (U), and Down (D). For notation convenience, we will refer to the number of modules in a Transformer block as $N_{mod}$. Thus, in LlaMA-2, $N_{mod} = 7$.

\noindent \textbf{LoRA} \quad For any Transformer module $m \in \text{\{Q, K, V, O, G, U, D\}}$, the LoRA method adds a pair of low-rank matrices to reparameterize its weights. Formally, the forward calculation of module $m$ with LoRA is: 
\begin{equation}
x^{'} = xW_{m} + xW_{m}^{A}W_{m}^{B} + b_{m}, 
\label{eq:lora_eq_1}
\end{equation}
where $W_{m} \in \mathbf{R}^{d_1 \times d_2 }$ is the weight matrix of module $m$, $b_{m}$ is its bias term. $W_{m}^{A} \in \mathbb{R}^{d_{1} \times r }$ and $W_{m}^{B} \in \mathbb{R}^{ r \times d_{2} }$ are the low-rank matrices for the LoRA module, and $r \ll \min (d_{1}, d_{2})$. $r$ is the rank of the two matrices and will also be referred to as the rank of the LoRA module.

\subsection{Motivation}

As demonstrated later in Table \ref{tab:results_efficiency_analysis}, the existing works on MOE style LoRA significantly slow down the LLM backbone during inference, reducing tokens per second (tps) by around 20\%. Each LoRA module is decomposed into multiple experts in these works, and a router should be called to determine which experts are activated. The calculations of multiple LoRA modules and multiple routers per layer are executed when generating every new token, resulting in latency that is not negligible. In order to improve the efficiency of such MOE LoRA methods, we need to investigate the following research questions: 

\noindent\emph{\textbf{RQ1.} Can we treat a LoRA module as an expert so that each Transformer layer has only one LoRA router and activate only one such expert per layer? }

\noindent\emph{\textbf{RQ2.} Can the LoRA router be called once for an input prompt? }

\subsection{Prompt-aware LoRA router} 

Trying to investigate \emph{\textbf{RQ1}} and \emph{\textbf{RQ2}}, we now try to propose the details of our MiLoRA method. The core of MiLoRA is the prompt-aware routing mechanism. Under this mechanism, the LoRA router takes the input prompt's hidden states as input and outputs the activated LoRA experts for the current layer. Different from the previous works \cite{chen2024llava,yang2024moral,liu2023moelora,dou2023loramoe,gou2023mixture}, our work: (a) only calculates the LoRA routers once when the input prompt is fed through the Transformer backbone for the first time and right before generating the first new token. The routers' activation decisions will be repeatedly used in the subsequent generation steps. (b) determine the activated LoRA experts at the Transformer's layer level, selecting which Transformer module is modified by its corresponding LoRA module.

As shown in Figure \ref{fig:architecture}, to generate a response, the input prompt has to go through the LLM backbone to obtain the hidden representations. Denote the hidden state of the input prompt with length $n_{p}$ right before Transformer layer $l$ as $\mathbf{H}^{l} \in \mathbf{R}^{n_{p} \times d}$. Then a pooling operation $\text{Pooler}()$ aggregates the semantic information in $\mathbf{H}^{l}$ and transforms it to $\mathbf{h}^{l} \in \mathbf{R}^{1 \times d}$:
\begin{equation}
\mathbf{h}^{l} = \text{Pooler}( \mathbf{H}^{l} ).
\end{equation}
Here, according to \cite{zhu-2021-autorc,Zhu2021AutoNLUAS}, the $\text{Pooler}$ operation can be one of the following: (a) last-token pooling, which is to use the vector representation of the last token in the prompt as $\mathbf{h}^{l}$. This pooler is widely used when decoder-based models perform sentence classification tasks. (b) average pooling. (c) max pooling. (d) self-attention-based pooling, whose detail is introduced in Appendix \ref{sec:self_attn_pooler}.

Then, $\mathbf{h}^{l}$ will go through an activation function $g$ and then the LoRA router $R^{l}$ right before layer $l$. $R^{l}$ assigns the current input prompt to the most suitable LoRA expert. This router contains (a) a linear layer that computes the probability of $\mathbf{h}^{l}$ being routed to each LoRA expert $\text{LoRA}_{m}$, (b) a softmax function to model a probability distribution over the LoRA experts, and finally, (c) a $\text{Top-k}$ function that choose the top $k > 0$ experts with the highest probability masses. Formally, 
\begin{equation}
R^{l}(\mathbf{h}^{l}) = \text{Top-k}( \text{Softmax} ( g(\mathbf{h}^{l}) W_{r}^{l} ) ), 
\label{eq:lora_router}
\end{equation}
where $W_{r}^{l} \in \mathbf{R}^{d \times N_{mod}}$ is the router's weight. The LoRA router dynamically selects the best $k$ experts for each input prompt during inference. Note that the router is only called once before a new token is generated. The activated LoRA experts are used throughout the whole generation process.

Following \citet{fedus2022switch}, we add a load balancing loss to the training loss function. Consider a training batch $B$ with $N_{B}$ samples, let $f_{i}^{l}$ represent the proportion of prompts assigned to the $i$-th LoRA expert in layer $l$, 
\begin{equation}
f_{i}^{l} = \dfrac{1}{N_{B}} \sum_{x\in B} \mathbf{1} \{ \arg\max_{j} p^{l}_{j}(x) = i \},
\end{equation}
where $p^{l}_{j}$ is the probability of expert $j$, output by the router $l$. Let $\hat{p}^{l}_{i}$ be the average of probability masses received by the $i$-th expert, $\hat{p}^{l}_{i} = \dfrac{1}{N_{B}} \sum_{x\in B} p^{l}_{i}(x) $. Then, the load balancing loss is given by:
\begin{equation}
\mathcal{L}_{lb} = N_{mod} \sum_{i=1}^{ N_{mod} } f_{i}^{l} \cdot \hat{p}^{l}_{i}.
\end{equation}
The $\mathcal{L}_{lb}$ loss term is added to the cross entropy loss with a coefficient $\lambda_{lb} \geq 0$.

\subsection{Learned activation functions}

The previous PEFT literature usually set the activation functions in a PEFT module to be ReLU \cite{Mahabadi2021CompacterEL,pfeiffer-etal-2021-adapterfusion,Liu2022LatePT} and does not discuss whether this setting is optimal. In addition, the PEFT modules' activation functions in different Transformer layers are usually set to be identical. As will be presented later in Table \ref{tab:ablations}, it is beneficial for LoRA routers of different depths to have different activation functions. Thus, how can we find an optimal setting for the LoRA routers' activation functions? Exhaustive hyper-parameter search is time and GPU-consuming. Thus, we are motivated to set the activation function to be learnable during training.

We resort to rational activation functions \cite{Molina2019PadAU}, which are learnable and can approximate common activation functions and learn new ones. The rational activation function
$R(x)$ of order $m$, $n$ is defined as follows: 
\begin{equation}
\text{Ra}(x) = \dfrac{ \sum_{j=0}^{m} a_{j} x^{j} }{  1 + \| 
\sum_{i=1}^{n} b_{i} x^{i} \| },   
\end{equation}
where $a_{j}$ and $b_{i}$ are learnable parameters. The rational activation functions are successfully applied in image classification \cite{Molina2019PadAU} and sequence modeling \cite{delfosse2021recurrent}.

Inspired by the above literature, we propose learning the activation functions in LoRA routers via the rational activation functions when finetuning a downstream task. Denote the set of parameters in the learnable activations as $\Theta$ and the other parameters in the LoRA routers and LoRA experts as $\Omega$. Following DARTS \cite{Liu2019DARTSDA}, we consider $\Theta$ as architectural parameters and optimize them along with $\Omega$ via bi-level optimization. Due to limited length, we introduce bi-level optimization in Appendix \ref{sec:appendix_bi_level_opt}.

\section{Experiments}

In this section, we conduct a series of experiments and analysis to evaluate our MiLoRA method.

\subsection{Datasets and evaluation metrics}

We compare our approach to the baselines on a collection of challenging tasks: (a) five benchmark common-sense question-answering tasks, ARC-e and ARC-c \cite{clark2018think}, OBQA \cite{mihaylov2018can}, PIQA \cite{bisk2020piqa}, BoolQ \cite{clark2019boolq}. (b) two math reasoning tasks,  AQuA \cite{ling2017program} and  GSM8k \cite{cobbe2021training}. We utilize the chain-of-thought (COT) rationales for these samples provided by \citet{hu2023llm} for training on these math tasks. All rationales are generated through zero-shot CoT \cite{Wei2022ChainOT,kojima2022large} on GPT-3.5\footnote{\url{https://platform.openai.com/docs/models}}, but without undergoing any error filtering. (c) MT-Bench \cite{2023arXiv230605685Z}, MMLU \cite{hendrycks2020measuring}, and BBH \cite{suzgun2022challenging}. Since these tasks provide no training data, we utilize the Alpaca \cite{alpaca} dataset for instruction tuning. The detailed statistics, and evaluation metrics can be found in Appendix \ref{sec:appendix_datasets}.

\subsection{Baselines}

We compare our MiLoRA framework with the current SOTA PEFT baseline methods. 

\noindent\textbf{LoRA and its variants} \ we consider the following LoRA variants as baselines: (a) the original LoRA \cite{hu2021lora}; (b) AdaLoRA \cite{Zhang2023AdaptiveBA}, which adaptively adjust the LoRA parameters among different Transformer modules. (c) MOELoRA \cite{liu2023moelora}, which considers each LoRA module as a mixture of single-rank LoRA experts. (d) DoRA \cite{liu2024dora}, one of the most recent variants of LoRA that decomposes the pre-trained weights into two components, magnitude, and direction, for fine-tuning, specifically employing LoRA for directional updates.

\noindent\textbf{Other PEFT methods} \ We also consider the most recent PEFT methods: (a) Parallel-Adapter proposed by \citet{He2021TowardsAU}; (b) Learned-Adapter \cite{Zhang2023LearnedAA}. (c) P-tuning v2 \cite{Liu2021PTuningVP}. (d) IAPT \cite{zhu2024iapt}. (e) BitFit \cite{BenZaken2021BitFitSP}. (f) (IA)$^{3}$ \cite{Liu2022FewShotPF}, which multiplies learnable vectors to the hidden states in different modules of the Transformer layer. (g) SSP \cite{Hu2022SparseSS}, which is a representative work on combining different PEFT methods, including LoRA and BitFit.

The baselines are implemented using their open-sourced codes. We only adjust the hyper-parameters related to tunable parameter numbers to fairly compare the baseline methods and our MiLoRA method.

\begin{table*}[tb!]
\centering
\resizebox{0.94\textwidth}{!}{
\renewcommand\arraystretch{1.1}
\begin{tabular}{c|cc|cccccccc}
\hline
\multirow{2}*{\textbf{Method}}   &   \textbf{Tunable}  &   \textbf{Activated}   &     \textbf{ARC-e}   &     \textbf{ARC-c}   &   \textbf{BoolQ}   &   \textbf{OBQA}  &  \textbf{PIQA}    &   \textbf{AQuA}   &    \textbf{GSM8k}    &   \multirow{2}*{\textbf{Avg.}}    \\ 

&  \textbf{Params}  &     \textbf{Params}   &    \textbf{(acc)}   &   \textbf{(acc)}     &  \textbf{(acc)}   &   \textbf{(acc)}  &   \textbf{(acc)}  &   \textbf{(acc)}   &   \textbf{(acc)}    &     \\
\hline

\multicolumn{9}{c}{\textbf{\emph{Baselines}}}  \\
\hline


Parallel-Adapter  &    83.9M   &     83.9M     &   67.1    &    54.2    &     65.2    &   76.3    &   69.8   &    15.6   &  26.4    &   53.5  \\
Learned-Adapter   &   81.8M   &    81.8M   &   69.3     &    54.4  &    64.9    &  78.4   &    75.6   &   18.3   &    28.9     &     55.7   \\
\hdashline
P-tuning v2    &    84.5M    &    84.5M    &    63.5   &     51.3    &  61.2    &  76.1    &  66.2    &   9.63    &  21.1   &    49.9   \\

IAPT    &    83.9M    &     83.9M    &   66.3    &     54.7    &  67.8    &  79.2    &    77.3     &   13.6     &   25.8     &    55.0     \\
\hdashline
BitFit &   87.0M   &   87.0M     &    65.9     &    54.1    &   66.4   &   77.2    &     76.6    &   11.8     &    21.7    &  53.4  \\
(IA)$^{3}$  &    78.6M   &     78.6M   &    68.1     &   54.6      &    67.2    &  78.1    &   75.4     &   13.2   &   23.4     &   54.3   \\
SSP &   80.6M   &    80.6M   &   71.6    
 &  57.6    &     69.6    &    79.5      &    79.7    &    15.9     &    31.8      &   58.0  \\

\hdashline

LoRA   &     80.0M   &   80.0M    &     73.4    &    57.2   &    68.8   &    80.1   &    81.4    &     16.6     &  31.1     &   58.4   \\ 

AdaLoRA   &  80.0M   &   80.0M    &    73.8    &    57.9   &    69.2   &    80.4   &    82.1    &  17.6    &   31.7     &   59.0 \\

MOELoRA &  87.3M    &   30.1M      &    
 76.8   &  60.2   &     72.0     &    81.1    &     82.7    &     18.3      &  32.3     &   60.4   \\
 
DoRA   &    80.0M   &   80.0M    &    76.5   &   59.8   &   71.7  &    80.6   &    82.7  
 &    17.9    &   32.6      &  60.3   \\

\hline
\multicolumn{9}{c}{\textbf{\emph{Our proposed methods}}}  \\
\hline

MiLoRA (ours)   &   80.9M    &     25.2M     &   \textbf{77.8}   &  \underline{61.2}   &  \underline{72.8}  &  \textbf{81.7}  &  \textbf{83.3}   &   \textbf{19.9}    &    \underline{33.9}     &    \textbf{61.5}    \\

MiDoRA (ours)   &   80.9M    &     25.8M     &   \underline{77.5}     &    \textbf{61.3}     &   \textbf{72.9}    &     \underline{81.3}    &   \underline{83.1}    &   \underline{19.3}    &       \textbf{34.1}     &    \underline{61.3}    \\

\hline
\end{tabular}}

\caption{\label{tab:results_main_1} The Overall comparison of different PEFT methods for single-task learning. The backbone model is LlaMA-2 7B. We report the median accuracy over five random seeds. Bold and Underline indicate the best and the second-best results.} 
\end{table*}

\subsection{Experiment Settings}
\label{subsec:experimental_settings}

\noindent\textbf{Computing infrastures} \quad We run all our experiments on NVIDIA A40 (48GB) GPUs. 

\noindent\textbf{Pretrained backbones} \quad The main experiments use the most recent open-sourced LLMs, LlaMA-2 7B \cite{Touvron2023Llama2O} as the pretrained backbone model. In the ablation studies, we will also use the recently released LlaMA-2 13B and Gemma 2B \cite{team2024gemma}. 

\noindent\textbf{Prediction heads} \quad When fine-tuning LlaMA-2 7B, we only consider the supervised fine-tuning (SFT) setting \cite{ouyang2022training}. After receiving a prompt or instruction, all the predictions are generated using the language modeling head (LM head). No additional prediction heads are installed to make categorical or numerical predictions. For decoding during inference, we use beam search with beam size 3.

\noindent\textbf{Hyper-parameters for the MiLoRA framework} \quad In our experiments, unless otherwise specified, we set: (a) the rank of each LoRA expert is set to $r=32$. (b) $k$ is set to 3. That is, each router activates one expert. (c) the LoRA router adopts the self-attention pooler. (d) the hyper-parameters of the rational activation are $m = 6$, $n=5$, and th
e learnable parameters $a_j$ and $b_i$ are initialized by approximating the GeLU activation function. (e) $\lambda_{lb}$ is set to 1e-2. Under the above settings, our MiLoRA method will introduce 80.9M tunable parameters and, at most, 16.4M activated PEFT parameters to the LlaMA-2 7B backbone. The hyper-parameters for training are specified in Appendix \ref{sec:appendix_exp_settings}.











\noindent\textbf{Reproducibility} \quad We run each task under five different random seeds and report the median performance on the test set of each task. 

Due to limited length, other experimental settings for the baseline methods and the training procedure are in Appendix \ref{sec:appendix_exp_settings}.

\subsection{Main results}
\label{subsec:main_results}

\noindent \textbf{Single-task setup.} \quad In this setup, We compare MiLoRA with baseline PEFT methods by employing these methods for fine-tuning a single task. The experimental results on the five commonsense reasoning tasks and two math reasoning tasks are presented in Table \ref{tab:results_main_1}. We present the number of tunable parameters in the second column and the average activated parameters in the third column. Table \ref{tab:results_main_1} reveals that our MiLoRA method outperforms the baseline methods across all seven tasks, with comparable tunable parameters and much fewer activated parameters. In particular, MiLoRA outperforms the previous SOTA LoRA style baselines like AdaLoRA, DoRA, and MOELoRA with comparable parameters. These results demonstrate that our method is good at downstream task adaptation of large language models.

\noindent \textbf{Multi-task setup.} \quad Table \ref{tab:results_main_multi_task} presents the results of LoRA, DoRA, MOELORA, and MiLoRA with LLaMA2-7B in multi-task learning. In contrast to the single-task setup in Table \ref{tab:results_main_1}, during multi-task learning, we mixed training data from ARC, BoolQ, OBQA, and PIQA to train the model, followed by separate evaluations to investigate the generalization ability of each method.
The results indicate that (a) compared to single-task learning, LoRA and DoRA exhibit degradation in average accuracy in multi-task learning (LoRA: -2.0\%, DoRA: -2.25\%). At the same time, MOELORA and MiLoRA maintain nearly the same average accuracy. MiLoRA presents nearly no performance loss regarding the average score.

\begin{table*}[tb!]
\centering
\resizebox{0.88\textwidth}{!}{
\renewcommand\arraystretch{1.1}
\begin{tabular}{c|cc|cccccc}
\hline
\multirow{2}*{\textbf{Method}}   &   \textbf{Activated}   &   \multirow{2}*{\textbf{ST/MT} }    &     \textbf{ARC-e}   &     \textbf{ARC-c}   &   \textbf{BoolQ}   &   \textbf{OBQA}  &  \textbf{PIQA}    &   \multirow{2}*{\textbf{Avg.}}    \\ 

&  \textbf{Params}   &     &    \textbf{(acc)}   &   \textbf{(acc)}     &  \textbf{(acc)}   &   \textbf{(acc)}  &   \textbf{(acc)}     &     \\
\hline

\multirow{2}*{LoRA }  &     \multirow{2}*{80.0M }   &  ST   &     73.4    &    57.2   &    68.8   &    80.1   &    81.4       &   72.2   \\    

&     &   MT     &      67.2 ( \textcolor{red}{-6.2} )    &  55.1 ( \textcolor{red}{-2.1} )  &   69.1  ( \textcolor{red}{+0.3} )  &  80.9  ( \textcolor{red}{+0.8} )    &   78.6   ( \textcolor{red}{-2.8} )     &   70.2 ( \textcolor{red}{-2.0} )      \\

\hdashline

\multirow{2}*{MOELoRA}  &    \multirow{2}*{17.3M }    &    ST    &      76.8   &  60.2   &     72.0     &    81.1    &     82.7    &    74.6    \\

&     &   MT     &    76.1   ( \textcolor{red}{-0.7} )     &  59.3   ( \textcolor{red}{-0.9} )      &   71.5 ( \textcolor{red}{+0.1} )   &  80.7 ( \textcolor{red}{-0.4} )      &   82.1  ( \textcolor{red}{-0.3} )      &    73.9   ( \textcolor{red}{-0.5} )   \\

\hdashline

\multirow{2}*{DoRA}   &    \multirow{2}*{80.0M}   &   ST   &    76.5   &   59.8   &   71.7  &    80.6   &    82.7    &     74.3   \\

&     &   MT     &   74.1  ( \textcolor{red}{-2.4} )     &  59.6 ( \textcolor{red}{-0.2} )  &   67.4 ( \textcolor{red}{-4.3} )  &     79.2 ( \textcolor{red}{-1.4} )    &   80.4 ( \textcolor{red}{-2.3} )   &      72.1 ( \textcolor{red}{-2.2} )     \\

\hdashline

\hline

\multirow{2}*{MiLoRA (ours) }  &    12.1M     &    ST   &   77.8   &  61.2   &     72.8   &      81.7     &  83.3    &    75.4       \\

&  12.3M   &   MT     &   77.4 ( \textcolor{red}{-0.4} )    &  61.5 ( \textcolor{red}{+0.3} )   &  72.3 ( \textcolor{red}{-0.3} )   &  81.3 ( \textcolor{red}{-0.4} )    &  83.5 ( \textcolor{red}{+0.3} )    &       75.2  ( \textcolor{red}{-0.1} )      \\

\hline
\end{tabular}}

\caption{\label{tab:results_main_multi_task} The Overall comparison of different PEFT methods for multi-task learning. The backbone model is LlaMA-2 7B. ST refers to the single-task setup, while MT refers to the multi-task setup. We report the average accuracy scores over five different runs, with the difference between MT and ST in red font in the brackets. } 
\end{table*}

\noindent \textbf{Results for general-purpose instruction tuning.} \quad After the LlaMA-2 7B is fine-tuned on the Alpaca \cite{alpaca} dataset with our MiLoRA method or the MOELoRA methods, we utilize the challenging benchmarks, MT-Bench \cite{2023arXiv230605685Z}, MMLU \cite{hendrycks2020measuring}, and BBH \cite{suzgun2022challenging}, for evaluation. We report the average GPT-4 score (gpt4-score) on the MT-Bench. Table \ref{tab:results_alpaca} presents the results. Consistent with the previous experiments (Table \ref{tab:results_main_1} and \ref{tab:results_main_multi_task}), our MiLoRA method outperforms the MOELoRA methods on the three benchmarks, demonstrating that MiLoRA is superior in enhancing the instruction tuning quality of large language models.

\begin{table}[tb!]
\centering
\resizebox{0.42\textwidth}{!}{
\renewcommand\arraystretch{1.2}
\begin{tabular}{c|ccc}
\hline
\multirow{2}*{\textbf{Method} }  &    \textbf{MT-Bench}     &    \textbf{MMLU}    &  
 \textbf{BBH}   \\ 

&   \textbf{gpt4-score} ($\uparrow$)   &    \textbf{acc}   &    \textbf{acc}    \\ 
\hline
MOELoRA    &   7.08   &    48.2    &  36.8   \\
\hdashline
MiLoRA   &   7.21   &   49.7   &   37.3   \\
\hline

\end{tabular}}
\caption{\label{tab:results_alpaca} Performance of general-purpose instruction tuning using the MiLoRA and MOELoRA methods. The backbone model is LlaMA-2 7B. $\uparrow$ means the metric is higher the better. }
\end{table}

\subsection{Ablation studies and further analysis}
\label{subsec:ablation_studies}

\noindent\textbf{Analysis of the inference efficiency} \quad To demonstrate the inference efficiency of our MiLoRA method, we now compare the GPU memory and decoding speed of MiLoRA, DoRA, and MOELoRA under beam search with different beam sizes. In this experiment, LoRA parameters are not merged to the backbone to mimic the single-LLM multi-tenant setting \cite{Chen2023PunicaML}. We present two metrics for measuring efficiency: (a) peak memory cost (in MiB). (b) tokens generated per second (tps). The results are presented in Table \ref{tab:results_efficiency_analysis}.

\begin{table}[tb!]
\centering
\resizebox{0.48\textwidth}{!}{
\renewcommand\arraystretch{1.05}
\begin{tabular}{c|ccc}
\hline
\multirow{2}*{\textbf{Method}}   &    \multirow{2}*{\textbf{Beam size}}  &  \textbf{Speed }   &   \textbf{Memory cost }     \\ 
&     &    \textbf{(tps)}    &   \textbf{(MiB)}  \\

\hline

\multirow{ 2}{*}{ DoRA }   &   1    &   36.5     &    13784    \\
        &   3   &    29.6    &   15292    \\

\hdashline
\multirow{ 2}{*}{ MOELoRA }   &   1   &   35.9    &  13788   \\
    &   3   &    28.4    &  15352    \\

\hdashline
\multirow{ 2}{*}{ MiLoRA }   &   1   &     43.7  &   13784     \\
    &   3   &   33.5   &   15300    \\
    
\hline
\end{tabular}}
\caption{\label{tab:results_efficiency_analysis} The memory and speed of LlaMA-2 7B for generating responses given input instructions, with different PEFT methods.  }
\end{table}

From Table \ref{tab:results_efficiency_analysis}, under beam sizes 1 and 3, the MiLoRA method has a comparable memory cost with MOELoRA and DoRA. However, its generation speed in terms of tps is significantly higher. With beam size 1, MiLoRA is 21.7\% faster than MOELoRA and 19.7\% faster than DoRA. With beam size 3, MiLoRA is 17.9\% faster than MOELoRA and 13.2\% faster than DoRA. The speed advantages of MiLoRA come from the following factors: (a) our method only calls the LoRA router at each Transformer layer when the input prompt goes through the LLM for the first time and right before generating the first new token. In contrast, MOELoRA and almost all the existing MOE-based LoRA variants require one to call multiple routers per layer when generating every new token. (b) our method significantly reduces the number of LoRA modules activated to modify the LLM backbone at each decoding step, making generating new tokens more efficient.

\begin{figure}
\centering
\includegraphics[width=0.48\textwidth]{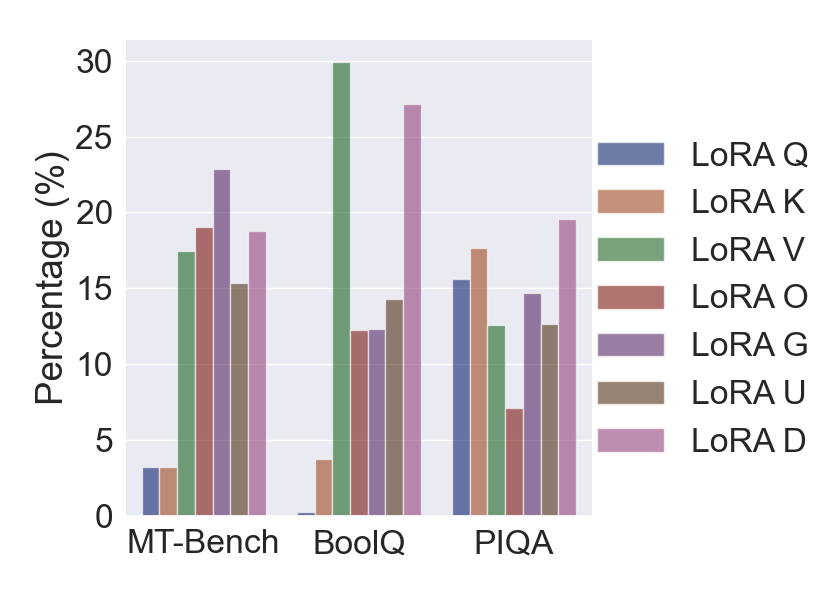}
\caption{Distribution of LoRA experts across Transformer layers. }
\label{fig:lora_dist}
\end{figure}

\noindent\textbf{Distributions of activated LoRA experts} \quad We now compare the distribution of LoRA experts across all Transformer layers on the MT-Bench, BoolQ, and PIQA tasks, in Figure \ref{fig:lora_dist}. We can observe that: (a) Different Transformer layers choose to activate different LoRA experts via their corresponding routers, and the maximum proportion a LoRA expert can achieve is less than 30\%. The results are intuitive since Transformer layers of different depths represent different knowledge, requiring different LoRA experts to express. (b) the LoRA distributions on different tasks are different. For example, a few layers activate LoRA Q or LoRA K on the MT-Bench and BoolQ tasks, while these two LoRA experts are frequently selected for the PIQA task.

\noindent\textbf{Ablation study of MiLoRA framework} \quad We now consider the following variants of MiLoRA: (a) MiLoRA-1 substitutes the self-attention pooling to average pooling. (b) MiLoRA-2 substitutes the self-attention pooling to the last-token pooling. (c) MiLoRA-3 uses the GeLU activation function $g$ for the LoRA router. (d) MiLoRA-4 uses ReLU for the first 16 layers' LoRA routers and GeLU for the deeper 16 layers'. (e) MiLoRA-5 uses GeLU for the first 16 layers' LoRA routers and ReLU for the deeper 16 layers'. The experimental results on the BoolQ, PIQA, and MMLU tasks are reported in Table \ref{tab:ablations}.

\begin{table}[tb!]
\centering
\resizebox{0.38\textwidth}{!}{
\begin{tabular}{c|ccc}
\hline
\multirow{2}*{\textbf{Method}}    &     \textbf{BoolQ}     &   \textbf{PIQA}   &    \textbf{MMLU}  \\ 

&    \textbf{(acc)}  &   \textbf{(acc)}   &   \textbf{(acc)}  \\
\hline
MiLoRA   &     \textbf{72.8}    &    \textbf{83.3}   &      \textbf{49.7}     \\
\hdashline

MiLoRA-1   &    72.5     &   83.1    &  49.5 \\
MiLoRA-2   &    72.4    &  82.9    & 49.6 \\
MiLoRA-3   &    72.3   &  82.8  &  49.3    \\
MiLoRA-4   &    71.5   &  82.0  &  48.7   \\
MiLoRA-5   &    72.4   &  82.9  &  49.4   \\
\hline
\end{tabular}}

\caption{\label{tab:ablations} The comparison of MiLoRA's variants on the BoolQ, PIQA, and MMLU tasks. The backbone model is LlaMA-2 7B. } 
\end{table}

The results show that MiLoRA under the default settings (as in Table \ref{tab:results_main_1}) outperforms the five variants. In addition, (a) comparing MiLoRA-1 and MiLoRA-2 to MiLoRA shows that the self-attention poolers provide high-quality information aggregation, leading to proper LoRA expert selection. (b) Comparing MiLoRA-5 to MiLoRA-3 and MiLoRA-4 demonstrates that using different activation functions for different layers' routers leads to a performance boost. (c) However, MiLoRA outperforms MiLoRA-3, MiLoRA-4, and MiLoRA-5, demonstrating that learnable activation functions can fit a proper activation function for each LoRA router and enhance downstream adaptation capability.

\begin{figure*}
\centering
\subfigure{%
\includegraphics[width=0.3\textwidth]{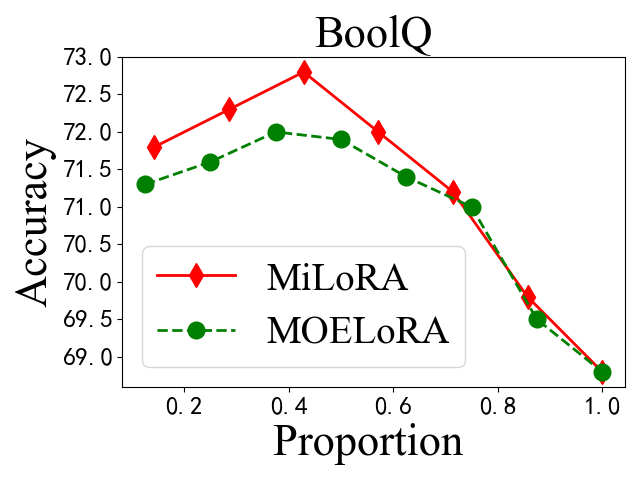}
\label{subfig:BoolQ_num_activated_experts}
}
\subfigure{%
\includegraphics[width=0.3\textwidth]{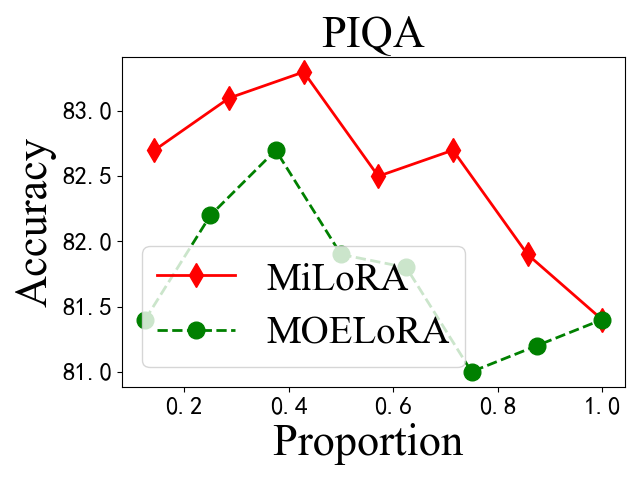}
\label{subfig:PIQA_num_activated_experts}
}
\caption{Performances under different proportion of activated experts. }
\label{fig:different_proportion_activated_experts}
\end{figure*}

\noindent\textbf{Effects of $k$.} \quad In Table \ref{tab:results_main_1} and \ref{tab:results_main_multi_task}, we set the number of activated LoRA experts, $k$, to 3. Now, we alter $k$ to \{1, 2, 4, 5, 6, 7\}, altering the proportion of activated LoRA experts. As a comparison, we also alter the proportion of activated experts in MOELoRA. The results of the BoolQ and PIQA tasks are presented in Figures \ref{subfig:BoolQ_num_activated_experts} and \ref{subfig:PIQA_num_activated_experts}, respectively. The results show that: (a) With the increased number of activated experts, the performance of the two methods first increases and then decreases. When the proportion of activated experts becomes 1, the two methods reduce to the vanilla LoRA. (b) Our MiLoRA consistently performs superior to the MOELoRA method, demonstrating our method's effectiveness in locating the Transformer modules that need LoRA modules the most.

\begin{figure*}
\centering
\subfigure{%
\includegraphics[width=0.3\textwidth]{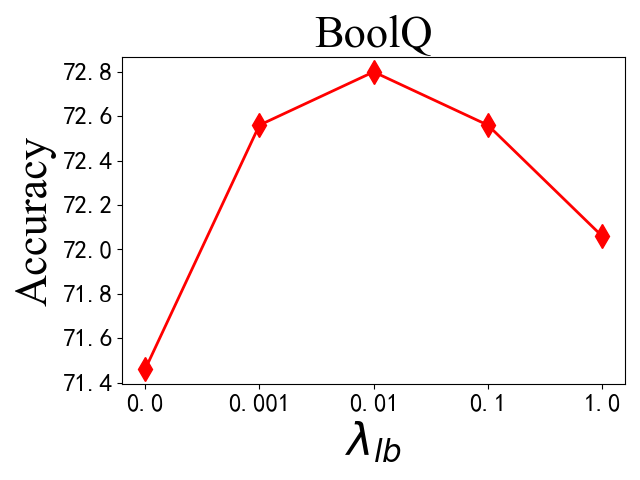}
\label{subfig:BoolQ_different_lambda_lb}
}
\subfigure{%
\includegraphics[width=0.3\textwidth]{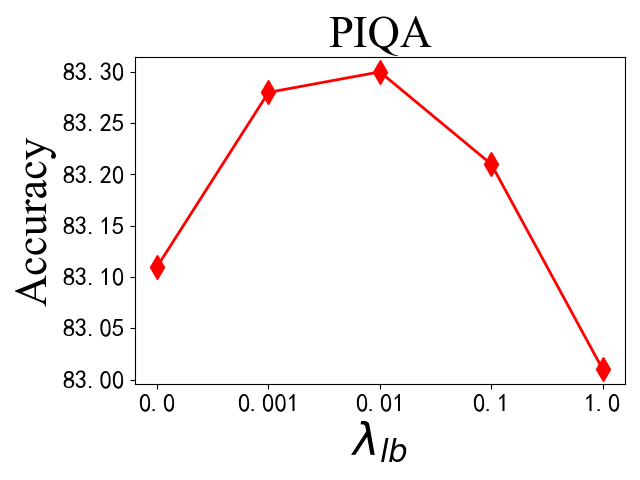}
\label{subfig:PIQA_different_lambda_lb}
}
\caption{Performances under different coefficient $\lambda_{lb}$. }
\label{fig:different_lambda_lb}
\end{figure*}

\noindent\textbf{Effects of the coefficient $\lambda_{lb}$} \quad In Table \ref{tab:results_main_1}, we set router loss coefficient, $\lambda_{lb}$, to 1e-2. Now, we alter $\lambda_{lb}$ to \{0.0, 1e-3, 1e-1, 1e0\}, and conduct experiments on the BoolQ and PIQA tasks. The results are reported in Figure \ref{subfig:BoolQ_different_lambda_lb} and \ref{subfig:PIQA_different_lambda_lb}. Results show that: (a) MiLoRA achieves the highest average accuracy with the coefficient 1e-2. (b) Disabling router loss or using a higher coefficient results in lower average accuracy. These results suggest that a reasonable router loss coefficient can help address the imbalance problem of experts,
while a higher coefficient can impede model convergence during fine-tuning.

\begin{figure*}
\centering
\subfigure{%
\includegraphics[width=0.3\textwidth]{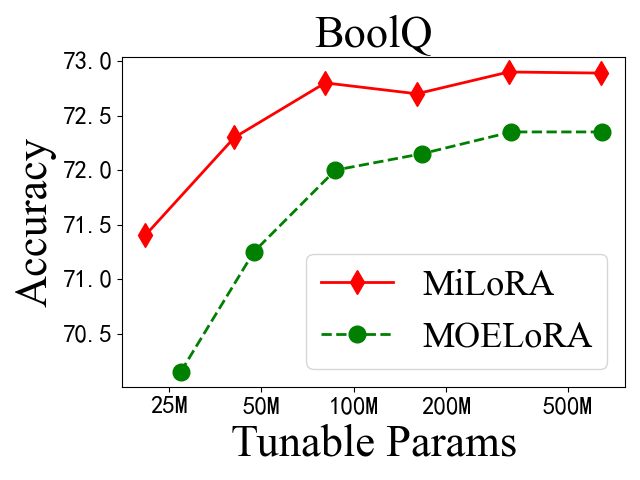}
\label{subfig:BoolQ_different_tunable_paras}
}
\subfigure{%
\includegraphics[width=0.3\textwidth]{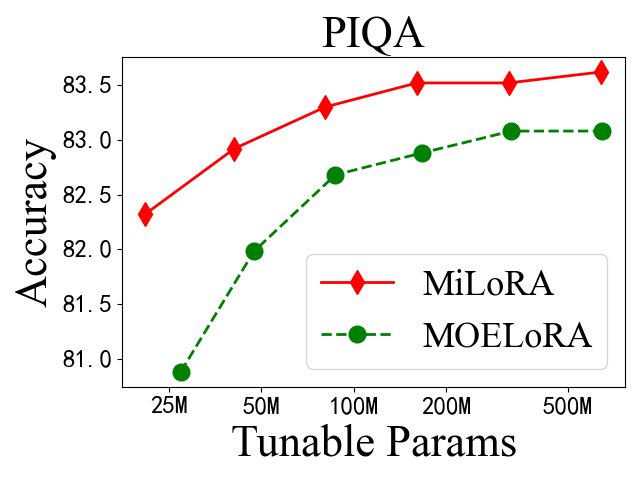}
\label{subfig:PIQA_different_tunable_paras}
}
\caption{Performances under different numbers of tunable parameters. }
\label{fig:different_tunable_parameters}
\end{figure*}

\noindent\textbf{Comparisons under different budgets of tunable parameters} \quad We vary the budget of tunable parameters for MiLoRA by modifying the values of $m=32$ to \{8, 16, 64, 128, 256\}. We also vary the MOELoRA method's tunable parameter numbers. The experimental results on the BoolQ and PIQA tasks are presented in Figure \ref{subfig:BoolQ_different_tunable_paras} and \ref{subfig:PIQA_different_tunable_paras}. The results show that under different tunable parameter budgets, our MiLoRA method (a) can consistently outperform the LoRA and LPT methods, and (b) is more robust to decreases in tunable parameter numbers.

\noindent\textbf{Ablation on the pretrained backbones} \quad Our main experiments are conducted on the LlaMA-2 7B model. To demonstrate the broad applicability of our method, we now conduct experiments on LlaMA-2 13B and Gemma 2B. The results are reported in Table \ref{tab:results_different_backbones} of Appendix \ref{sec:appendix_ablation_pretrained_backbones}. We can see that our MiLoRA method can also outperform the baseline methods on these two backbones.

\section{Conclusion}

This work presents the Mixture of LoRA (MiLoRA) method, a novel method for the parameter-efficient fine-tuning of large language models. Different from previous literature on MOE style LoRA methods, MiLoRA: (a) activates LoRA experts at the Transformer layer level, determining which Transformer module's LoRA is activated. (b) The decision to activate which LoRA expert is conditioned on the input prompt. (c) for a given prompt, the LoRA routers are called only once. The subsequent token generation steps reuse the routers' decisions. In order to improve our framework's downstream performance, we propose to learn different activation functions during fine-tuning for LoRA routers of different depths. Our method is convenient to implement and off-the-shelf. Experiments on various tasks demonstrate that our MiLoRA method outperforms the baseline methods while being efficient in inference.

\section*{Limitations}

We showed that our proposed method can improve the performance of parameter-efficient tuning on diverse tasks and different pretrained models (i.e., LlaMA-2 7B, LlaMA-2 13B, Gemma 2B). However, we acknowledge the following limitations: (a) the more super-sized open-sourced LLMs, such as LlaMA-2 70B, are not experimented due to limited computation resources. (b) Other tasks in natural language processing, like information extraction, were also not considered. But our framework can be easily transferred to other backbone architectures and different types of tasks. It would be of interest to investigate if the superiority of our method holds for other large-scaled backbone models and other types of tasks. And we will explore it in future work.

\section*{Ethics Statement}

The finding and proposed method aims to improve the soft prompt based tuning in terms of better downstream performances whiling pursuing efficiency. The used datasets are widely used in previous work and, to our knowledge, do not have any attached privacy or ethical issues. In this work, we have experimented with LlaMA-2 models, a modern large language model series. As with all LLMs, LlaMA-2’s potential outputs cannot be predicted in advance, and the model may in some instances produce inaccurate, biased or other objectionable responses to user prompts. However, this work's intent is to conduct research on different fine-tuning methods for LLMs, not building applications to general users. In the future, we would like to conduct further tests to see how our method affects the safety aspects of LLMs.

\bibliography{custom}
\bibliographystyle{acl_natbib}

\appendix

\section{Appendix: introduction to bi-level optimization}
\label{sec:appendix_bi_level_opt}

The bi-level optimization \cite{Liu2019DARTSDA} optimize $\Theta$ conditioned on the optimized parameters of $\Omega^{*}$. Denote the training set as $\mathcal{D}_{train}$, and the validation set as $\mathcal{D}_{val}$. The inner and outer levels of optimization are conducted on these two separate splits of the task dataset, which is analogous to validating architectures trained on $\mathcal{D}_{train}$ using a different split $\mathcal{D}_{val}$ to avoid over-fitting. Thus the optimization objective is:
\begin{align}
& \min_{\Theta} \mathcal{L}(\mathcal{D}_{val}, \Omega^{*}, \Theta),  \nonumber\\
& \emph{s.t.} \ \ \Omega^{*} =  \arg\min_{\Omega} \mathcal{L}(\mathcal{D}_{train}, \Omega, \Theta), 
\label{eq:bi_level_optimize}
\end{align}
where $\mathcal{L}()$ is the objective function on a given downstream task, such as cross entropy loss. The above bi-level optimization problem is approximated with an alternating optimization strategy. The gradients of $\Omega$ are calculated with batches of samples from $\mathcal{D}_{train}$, and the gradients of $\Theta$ are calculated on $\mathcal{D}_{val}$.

\section{Appendix for the datsets and evaluation metrics}
\label{sec:appendix_datasets}

\subsection{Dataset statistics}

The detailed statistics of the above tasks' datasets are presented in Table \ref{tab:dataset_stats}.

\begin{table*}[tb!]
\centering
\resizebox{0.8\textwidth}{!}{
\begin{tabular}{cccccc}
\hline
Datasets  &  \#train    &  \#dev   &   \#test   &     Type   &   Metrics  \\ 
\hline
\multicolumn{6}{c}{\textbf{\emph{Commonsense reasoning tasks}}}   \\
\hline
BoolQ  &     9427	  &  -  &   3270   &  Commonsense reasoning   &  acc    \\
OBQA    &  4957	   &   500	    &   500     &  Commonsense reasoning   &  acc      \\

ARC-e   &    2251	 &   570   &   2376    &  Commonsense reasoning   &  acc      \\
ARC-c   &    1119	 &  299	  &   1172    &  Commonsense reasoning   &  acc      \\

PIQA   &   16,000   &    2,000     &    3,000   &  Commonsense reasoning   &  acc \\

\hline
\multicolumn{6}{c}{\textbf{\emph{Math reasoning tasks}}}  \\
\hline
AQuA    &    97467	 &   254	 &   254   &   Math reasoning  &       acc          \\
GSM8K  &   7473   &   -	   &  1319   &   Math reasoning  &       acc    \\

\hline
\multicolumn{6}{c}{\textbf{\emph{Instruction tuning }}}  \\
\hline

Alpaca   &    50k  &    -    &  -   &  Instruction tuning  &  -     \\

\hline
\multicolumn{6}{c}{\textbf{\emph{LLM evaluation tasks}}}  \\
\hline

MT-Bench   &   -   &  -  &   80  &   Question answering  &  GPT-4 scores       \\

MMLU  &  -  &  -  &  14042      &   Question Answering    &   acc    \\

BBH  &  -  &  -  &  6,511   &     Question Answering    &    acc    \\

\hline
\end{tabular}}
\caption{\label{tab:dataset_stats}  The dataset statistics. }
\end{table*}

\subsection{Evaluation metrics/protocols}
\label{sec:appendix_evaluations}

For the commonsense reasoning and math reasoning tasks, since they usually come with a definite answer choice, we will directly consider the correctness of the final answers. Thus, we report accuracy (denoted as acc).

For evaluating the quality of instruction tuned LlaMA-2 7B on the MT-Bench, we follow the current common practice of utilizing GPT-4 as a unbiased reviewer \cite{2023arXiv230605685Z}. We generate model responses from a fine-tuned model with beam size 3 with the generation function in Huggingface Transformers \cite{wolf2020transformers}. Then we compare MOELoRA and MiLoRA's answers with GPT-4. For each instruction in MT-Bench, GPT-4 \cite{gpt4} is asked to write a review for both answers from the two methods, and assigns a quantitative score on a scale of 10 to each response.

\section{Details for the self-attention based pooler}
\label{sec:self_attn_pooler}

Our LoRA routers must pool the input prompts of variable lengths to a fixed length. For the pooling operation, the previous literature often chooses average pooling or max pooling \cite{Kim2014ConvolutionalNN,autotrans,Zhu2021AutoNLUAS}, which are pointed out by the literature \cite{zhu-2021-autorc} that they are prone to weaken important words when the input sequence is long, thus dropping useful information during pooling. Thus, in this work, we utilize the self-attention mechanism in our pooling module $\text{Pooler}()$. Self-Attention assigns each token in the input instruction a weight to indicate the importance of the token. A few crucial tokens to the task will be emphasized, while the less important tokens are ignored. Formally, we initialize a learnable weight matrix $W_{sa} \in \mathbb{R}^{d \times 1} $, then the self-attention based pooler's calculation processes are:
\begin{align}
\mathbf{U} & = \mathbf{h} W_{sa},  \nonumber \\   
\mathbf{A} & = \text{Softmax}(\mathbf{U}),  \nonumber \\  
\mathbf{p} & = \mathbf{A}^{\intercal} \mathbf{h},
\end{align}
where $\mathbf{p} \in \mathbb{R}^{n_{p} \times d} $ is the input tesor, $\text{Softmax}$ is the softmax function along the first dimension, and $\intercal$ denotes matrix transpose. In the above equations, each column of $W_{sa}$ is a trainable query vector designated to determine the self-attention weights via dot products between this query and each token. Then, the weights are normalized across the sequence dimension via the softmax normalization function. Corresponding to different soft tokens, different query vectors in $W_{sa}$ can aggregate the input instructions in different aspects, thus providing a high-quality summarization of the instruction's semantic information.










\section{Appendix for Experimental settings}
\label{sec:appendix_exp_settings}

Here, we provide more details for experimental settings. 

\noindent\textbf{Hyper-parameters for the baseline PEFT methods} \quad For P-tuning V2, the number of prompt tokens at each layer is set to 16, and the soft prompts are initialized with dimension 640, and then is projected to dimension 4096. For IAPT, the prompt length is 4, and the bottleneck dimension for the prompt generator is 320.

For the Parallel-Adapter and Learned-Adapter, the bottleneck dimension is set to 160. Adapters are connected to both the self-attention and FFN sub-layer. 

We adjust the sparsity for SSP so that the number of tunable parameters is comparable with MiLoRA and the other baselines. For BitFit, the bias vectors are initialized with dimension 64, and then a learnable projection layer projects it to the same dimension with the LlaMA-2 backbone. For (IA)$^{3}$, the activation adjusting vectors are added the Query, Key, and Up activations. The adjusting vectors are initialized with dimension 128, and then a learnable projection layer projects it to the same dimension with the LlaMA-2 backbone. 

For LoRA, the rank size $r$ at each LoRA module is set to 32. For AdaLoRA, the initial rank at each module is set to 64, and half of the rank budget is pruned during fine-tuning. For MOELoRA, the rank size $r$ at each LoRA module is set to 32, and the LoRA modules is reformulated as 32 single-rank LoRAs. Then each 4 forms an expert. Thus, a LoRA module consists of 8 experts, and the router is top-4 router, activating 4 of the expert for predicting the next token. DoRA also sets the rank size $r$ to 32.

\noindent\textbf{Training settings for PEFT methods} \quad We use the HugginFace Transformers \cite{wolf-etal-2020-transformers}, PEFT \cite{peft}, or the original code repositories for implementing all the methods, and for training and making predictions. For fine-tuning LlaMA-2 7B model, the maximum sequence length is set to 768. The maximum training epoch is set to 10. The batch size is set between 16 for task with less than 10k training set, and 128 otherwise. We use AdamW as the optimizer with a linear learning rate decay schedule and 6\% of the training steps for warm-up. The learning rate is set to 1e-4. For MiLoRA, the load balance loss coefficient $\lambda_{lb}$ is set to 1e-2. For the bi-level optimization of learnable activations, the validation set is the same with the dev set. The hyper-parameters for calculating the gradients of the architectural parameters are the same with the normal training procedure, except that the learning rate is 1e-6. The other hyper-parameters are kept the same with \cite{wolf-etal-2020-transformers}. In every 200 steps, the model is evaluated on the dev set to calculate dev set perplexity. Patience is set to 10, that is, if the model does not achieve a lower dev set perplexity for 10 evaluation runs, the training stops early. The best checkpoint on the dev set is used to run predictions on the test set.

\section{Ablation on the pretrained backbones}
\label{sec:appendix_ablation_pretrained_backbones}

Our main experiments are conducted on the LlaMA-2 7B model. To demonstrate that our method works well regardless of the backbone models, we now conduct experiments on the LlaMA-2 13B model and Gemma 2B models. The other experimental settings are kept the same with the main experiments (Table \ref{tab:results_main_1}). We conduct experiments on the BoolQ, PIQA and MMLU tasks. The results are reported in Table \ref{tab:results_different_backbones}.

\begin{table}[tb!]
\centering
\resizebox{0.38\textwidth}{!}{
\begin{tabular}{c|ccc}
\hline
\multirow{2}*{\textbf{Method}}    &     \textbf{BoolQ}     &   \textbf{PIQA}   &    \textbf{MMLU}     \\ 

&    \textbf{(acc)}  &   \textbf{(acc)}   &   \textbf{(acc)}     \\
\hline 

\multicolumn{4}{c}{\textbf{\emph{Results for LlaMA-2 13B}}}  \\
\hline
MOELoRA   &     73.5     &    85.8    &   50.5 \\
\hdashline
MiLoRA    &    74.9    &   86.6    &   51.2   \\

\hline 
\multicolumn{4}{c}{\textbf{\emph{Results for Gemma 2B}}}  \\
\hline
MOELoRA   &    62.3   &    79.4    &   39.8  \\
\hdashline
MiLoRA    &   63.9    &    80.3   &   40.7    \\

\hline

\end{tabular}}
\caption{\label{tab:results_different_backbones} Results for different PEFT methods on the BoolQ, PIQA and MMLU benchmarks. The backbone LMs are LlaMA-2 13B, an Gemma 2B.}
\end{table}

\end{CJK*}

\end{document}